# On Using Linear Diophantine Equations to Tune the extent of Look Ahead while Hiding Decision Tree Rules

Georgios Feretzakis and Dimitris Kalles and Vassilios S. Verykios

School of Science and Technology, Hellenic Open University, Patras, Greece
georgios.feretzakis@ac.eap.gr, kalles@eap.gr, verykios@eap.gr

## Abstract

This paper focuses on preserving the privacy of sensitive patterns when inducing decision trees. We adopt a record augmentation approach for hiding sensitive classification rules in binary datasets. Such a hiding methodology is preferred over other heuristic solutions like output perturbation or cryptographic techniques - which restrict the usability of the data - since the raw data itself is readily available for public use. In this paper, we propose a look ahead approach using linear Diophantine equations in order to add the appropriate number of instances while minimally disturbing the initial entropy of the nodes.

## 1 Introduction

Privacy preserving data mining (Verykios et al., 2004) is a quite recent research area trying to alleviate the problems stemming from the use of data mining algorithms to the privacy of the data subjects recorded in the data and the information or knowledge hidden in these piles of data. Agrawal and Srinkant (Agrawal and Srinkant, 2000) were the first to consider the induction of decision trees from anonymized data, which had been adequately corrupted with noise to survive from privacy attacks. The generic strand of knowledge hiding research (Gkoulalas-Divanis and Verykios, 2009) has led to specific algorithms for hiding classification rules, like, for example, noise addition by a data swapping process (Estivill-Castro and Brankovic, 1999).

A key target area concerns individual data privacy and aims to protect the individual integrity of database records to prevent the re-identification of individuals or characteristic groups of people from data inference attacks. Another key area is sensitive rule hiding, the subject of this paper, which deals with the protection of sensitive patterns that arise from the application of data mining techniques. Of course, all privacy preservation techniques strive to maintain data information quality.

The main representative of statistical approaches (Chang and Moskowitz, 1998) adopts a parsimonious downgrading technique to determine whether the loss of functionality associated with not downgrading the data, is worth the extra confidentiality. Reconstruction techniques involve the redesign of the public dataset (Natwichai et al., 2005; Natwichai et al., 2006) from the non-sensitive rules produced by algorithms like C4.5 (Quinlan, 1993) and RIPPER (Cohen, 1995). Perturbation based techniques involve the modification of transactions to support only non-sensitive rules (Katsarou et al., 2009), the removal of tuples associated with sensitive rules (Natwichai et al., 2008), the suppression of certain attribute values (Wang et al., 2005) and the redistribution of tuples supporting sensitive patterns so as to maintain the ordering of the rules (Delis et al., 2010).

In this paper, we propose a series of techniques to efficiently protect the disclosure of sensitive knowledge patterns in classification rule mining. We aim to hide sensitive rules without compromising the information value of the entire dataset. After an expert selects the sensitive rules, we modify class labels at the tree node corresponding to the tail of the sensitive pattern, to eliminate the gain attained by the information metric that caused the splitting. Then, we appropriately set the values of non-class attributes, adding new instances along the path to the root where required, to allow non-sensitive patterns to remain as unaffected as possible (Kalles et al., 2016). This approach is of great importance as the sanitized data set can be subsequently published and, even, shared with competitors of the data set owner, as can be the case with retail banking [Li et al., 2011]. In this paper, we extend a previous work (Kalles et al., 2016) by formulating a generic look ahead solution which takes into account the tree structure all the way from an affected leaf to the root. The rest of this paper is structured in 3 sections. Section 2 describes the dataset operations we employ to hide a rule while attempting to minimally affect the decision tree. Section 3 discusses further research issues and concludes the paper.

## 2 The Baseline Problem and a Heuristic Solution

Figure 1 shows a baseline problem, which assumes a binary decision tree representation, with binary-valued, symbolic attributes ($X$, $Y$ and $Z$) and binary classes ($C_1$ and $C_2$).
Hiding $R_3$ implies that the splitting in node $Z$ should be suppressed, hiding $R_2$ as well.

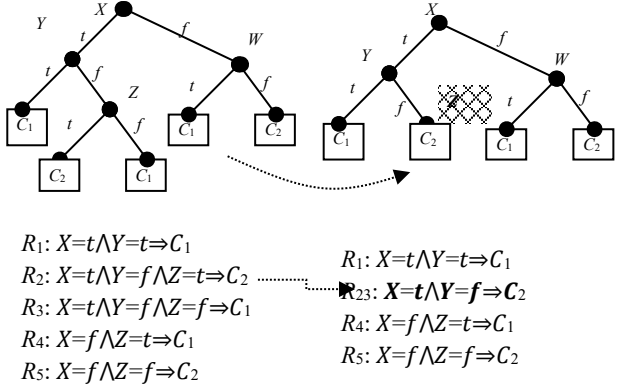

$R_1$: $X=t \wedge Y=t \Rightarrow C_1$
$R_2$: $X=t \wedge Y=f \wedge Z=t \Rightarrow C_2$ ········
$R_3$: $X=t \wedge Y=f \wedge Z=f \Rightarrow C_1$
$R_4$: $X=f \wedge Z=t \Rightarrow C_1$
$R_5$: $X=f \wedge Z=f \Rightarrow C_2$

$R_1$: $X=t \wedge Y=t \Rightarrow C_1$
$R_{23}$: $X=t \wedge Y=f \Rightarrow C_2$
$R_4$: $X=f \wedge Z=t \Rightarrow C_1$
$R_5$: $X=f \wedge Z=f \Rightarrow C_2$

*Figure 1. A binary decision tree before (left) and after (right) hiding and the associated rule sets.*

A first idea to hide $R_3$ would be to remove from the training data all the instances of the leaf corresponding to $R_3$ and to retrain the tree from the resulting (reduced) dataset. However this action may incur a substantial tree restructuring, affecting other parts of the tree too.

Another approach would be to turn into a new leaf the direct parent of the $R_3$ leaf. However, this would not modify the actual dataset, thus an adversary could recover the original tree.

To achieve hiding by modifying the original data set in a minimal way, we may interpret "minimal" in terms of changes in the data set or in terms of whether the *sanitized* decision tree produced via hiding is syntactically close to the original one. Measuring minimality in how one modifies decision trees has been studied in terms of heuristics that guarantee or approximate the impact of changes (Kalles and Morris, 1996; Kalles and Papagelis, 2000; Kalles and Papagelis, 2010).

However, hiding at $Z$ modifies the statistics along the path from $Z$ to the root. Since splitting along this path depends on these statistics, the relative ranking of the attributes may change, if we run the same induction algorithm on the modified data set. To avoid ending up with a completely different tree, we first employ a bottom-up pass (*Swap-and-Add*) to change the class label of instances at the leaves and then to add some new instances on the path to the root, to preserve the key statistics at the intermediate nodes.

Then, we employ a top-down pass (*Allocate-and-Set*) to complete the specification of the newly added instances. These two passes help us hide all sensitive rules and keep the sanitized tree close to the form of the original decision tree.

These two techniques had been fully described in previous published works (Kalles et al., 2016). The main contribution of this paper is the improvement of the Swap-and-Add pass by following a look ahead approach than a greedy which was used before.

### 2.1 Adding instances to preserve the class balance using Linear Diophantine Equations: a proof of concept and an indicative example

The *Swap-and-Add* pass aims to ensure that node statistics change without threatening class-value balances in the rest of the tree. Using Figure 2 as an example, we show the original tree with class distributions of instances across edges.

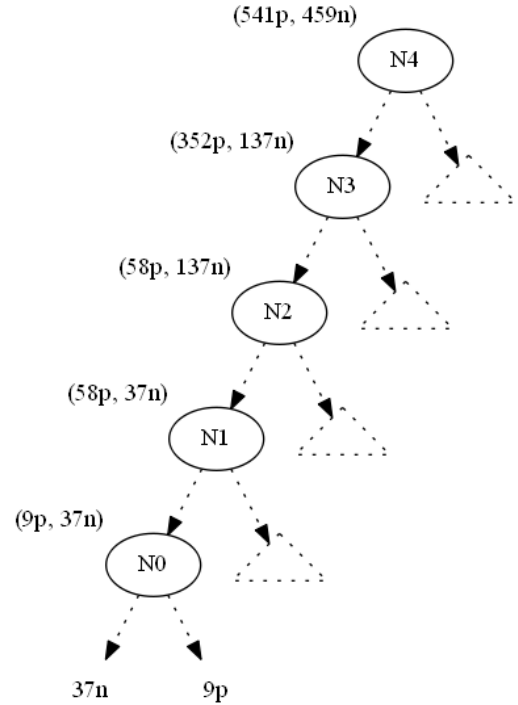

*Figure 2. Original tree*

We use the information gain as the splitting heuristic. To hide the leaf which corresponds to the 9 positive instances (to the right of N0) we change the nine positive instances to negative ones and denote this operation by (-9$p$,+9$n$). As a result the parent node, *N0*, becomes a one-class node with minimum (zero) entropy. All nodes located upwards to node

*N0* until the root *N4* also absorb the (-9p, +9n) operation (Figure 3).
This conversion would leave *N1* with 49p+46n instances. But, as its initial 58p+37n distribution contributed to *N1*'s splitting attribute, $A_{N1}$, which in turn created *N0* (and then 9p), we should preserve the information gain of $A_{N1}$, since the entropy of a node only depends on the ratio *p:n* of its instance classes (Lemma 1).

**Lemma 1**. *The entropy of a node only depends on the ratio of its instance classes.*
(The proof is in (Kalles et al., 2016))

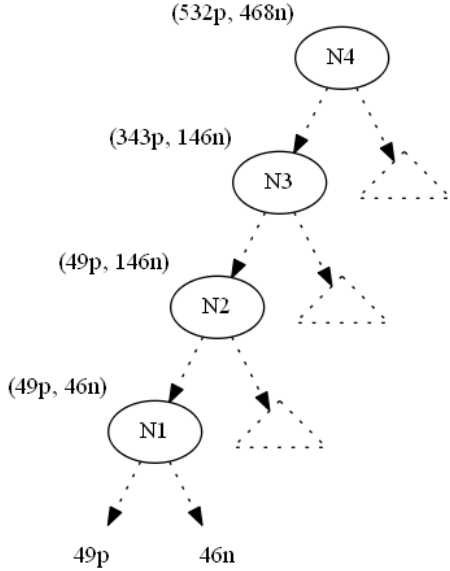

Figure 3. Bottom-up propagation of instances (-9p,+9n).

To maintain the initial ratio ($\frac{58p}{37n}$) of node N1, we should add appropriate number of positive and negative instances to N1 and extend this addition process up until the tree root, by accumulating at each node all instance requests from below and by adding instances locally to maintain the node statistics, propagating these changes to the tree root.
In a previously published work (Kalles et al., 2016) the above procedure was greedy, essentially solving the problem for only one (tree) level of nodes, which resulted many times in a non-optimum (minimum) number of added instances, whereas a look ahead based solution would be able to take into account all levels up to the root. In addition, the new ratios (p:n) of the nodes were not exactly the same as they were before the change, thus propagating ratio changes whose impact could only be quantified in a compound fashion by inspecting the final tree and hampering our ability to investigate the behavior of this heuristic in a detailed fashion. We, therefore, reverted to using Diophantine Linear Equations as the formulation technique of the problem of determining how many instances to add; as we shall show, this technique deals with both issues in one go.

Let $(x_1, y_1)$ be the number of positive and negative instances respectively that have to be added to node N1 in order to maintain its initial ratio. This can be expressed with the following equation:
$$\frac{49 + x_1}{46 + y_1} = \frac{58}{37}$$
The above equation is equivalent to the following linear Diophantine equation:

$$37x_1 - 58y_1 = 855 \qquad (1)$$

Similarly, let $(x_2, y_2), (x_3, y_3), (x_4, y_4)$ be the corresponding number of positive and negative instances that have to be added to nodes N2, N3 and N4.
The corresponding linear Diophantine equations for nodes N2, N3 and N4 are:

$$137x_2 - 58y_2 = 1755 \qquad (2)$$
$$137x_3 - 352y_3 = 4401 \qquad (3)$$
$$459x_4 - 541y_4 = 9000 \qquad (4)$$

The general solutions of the above four (1-4) linear Diophantine equations are given below (k ∈ ℤ):

$$37x_1 - 58y_1 = 855 \Leftrightarrow \begin{cases} x_1 = 9405 + 58k \\ y_1 = 5985 + 37k \end{cases}$$
$$137x_2 - 58y_2 = 1755 \Leftrightarrow \begin{cases} x_2 = -19305 + 58k \\ y_2 = -45630 + 137k \end{cases}$$
$$137x_3 - 352y_3 = 4401 \Leftrightarrow \begin{cases} x_3 = -734967 + 352k \\ y_3 = -286065 + 137k \end{cases}$$
$$459x_4 - 541y_4 = 9000 \Leftrightarrow \begin{cases} x_4 = -297000 + 541k \\ y_4 = -252000 + 459k \end{cases}$$

From the infinite pairs of solutions for every linear Diophantine equation we choose the pairs
$(x_1^*, y_1^*), (x_2^*, y_2^*), (x_3^*, y_3^*)$ and $(x_4^*, y_4^*)$, where $x_1^*, x_2^*, x_3^*, x_4^*$, $y_1^*, y_2^*, y_3^*, y_4^*$ are the minimum natural numbers that satisfy the following condition.
(C1): $x_1^* \leq x_2^* \leq x_3^* \leq x_4^*$ and $y_1^* \leq y_2^* \leq y_3^* \leq y_4^*$
Condition (C1) ensures that we have selected the optimum path to the root of the decision tree in terms that every addition of instances propagates upwards in a consistent manner (i.e. if one adds some instances at a lower node, one cannot have added fewer instances in an ancestor node).
With this technique we can determine exactly the minimum number of instances that must be added to each node in order to maintain the initial ratios of every node.
There are a few cases that a linear Diophantine equations has no solutions, this problem can be overcome by a little change to the initial ratio until we construct a solvable linear

Diophantine equation. For this example the pairs of solutions that are both minimum and satisfy the condition (C1) are:
$(x_1^*, y_1^*) = (67, 28)$
$(x_2^*, y_2^*) = (67, 128)$
$(x_3^*, y_3^*) = (361, 128)$
$(x_4^*, y_4^*) = (550, 450)$

Based on the above solutions we have to add to N1, 67 positive and 28 negative instances which leads to a ratio of $(\frac{116p}{74n})$. These new instances propagate upwards, therefore on N2, we don't need to add any positive instances but we need to add 100 (=128-28) negative instances which leads to a ratio of $(\frac{116p}{274n})$. Similarly, for N3 we should add, 294 (=361-67) new positive and no negative instances. Finally, for N4, we should add 189 (=550-361) new positive and 322 (=450-128) new negative instances. Therefore, with this look ahead technique we know from the very beginning which one is the optimum path in order to add appropriate number of instances to maintain the exact values of initial ratios which means that we will not have any disturbance in our tree after the hiding.

We observe that the solutions of Diophantine equation (4) which corresponds to node N4 (root) determine the total number of instances that should be added to our dataset in order to have the same ratios as initially. If we slightly change the ratio of N4 (in our case let be changed to $(\frac{540p}{460n})$ instead of $(\frac{541p}{459n})$) then we will have a different Diophantine equation which leads to a smaller number of added instances. In our example the new Diophantine equation (4') and the set of solutions are given below:

$$460x_4 - 540y_4 = 4000 \qquad (4')$$

$$460x_4 - 540y_4 = 4000 \Leftrightarrow \begin{cases} x_4 = -1400 + 27k \\ y_4 = -1200 + 23k \end{cases}, k \in \mathbb{Z}$$

For this example the pairs of solutions that are both minimum and satisfy the condition C1 are:
$(x_1^*, y_1^*) = (67, 28)$
$(x_2^*, y_2^*) = (67, 128)$
$(x_3^*, y_3^*) = (361, 128)$
$(x_4^*, y_4^*) = (382, 318)$

Therefore, we have to add 700 new instances instead of 1000 that we had to add before. Of course now we don't have exactly the same ratio p:n for node N4 but something very close to it. In other words, the method of linear Diophantine equations helping us to make a trade-off between the number of added instances and the accuracy of a node's ratio.

## 2.2 Fully specifying instances

Having set the values of some attributes for the newly added instances is only a partial instance specification, since we have not set those instance values for any other attribute other than the ones present in the path from the root to the node where the instance addition took place. Unspecified values must be so set to ensure that currently selected attributes at all nodes do not get displaced by competing attributes. This is what the *Allocate-and-Set* pass does.

With reference to Figure 2 and the $9n$ instances added due to N1 via the N2-N1 branch, these instances have not had their values set for $A_{N1}$ and $A_{N2}$. Moreover, these must be so set to minimize the possibility that $A_{N2}$ is displaced from N2, since (at N2) any of attributes $A_{N0}$, $A_{N1}$ or $A_{N2}$ (or any other) can be selected. Those $9n$ instances were added to help guarantee the existence of N1.

As it happened in the bottom-up pass, we need the information gain of $A_{N2}$ to be large enough to fend off competition from $A_{N0}$ or $A_{N1}$ at node N2, but not too large to threaten $A_{N3}$. We start with the best possible allocation of values to attribute $A_{N2}$, and progressively explore directing some of these along the N2 N1 branch, and stop when the information gain for $A_{N2}$ becomes lower than the information gain for $A_{N3}$. We use the term *two-level hold-back* to refer to this technique, as it spans two tree levels. This approach exploits the convexity property of the information gain difference function (Lemma 2).

The *Allocate-and-Set* pass examines all four combinations of distributing all positive and all negative instances to one branch, select the one that maximizes the information gain difference and then move along the slope that decreases the information gain, until we do not exceed the information gain of the parent; then perform the recursive specification all the way to the tree fringe.

**Lemma 2.** *Distributing new class instances along only one branch maximizes information gain.*
(The proof is in (Kalles et al., 2016))

## 2.3 Grouping of hiding requests

By serially processing hiding requests, each one incurs the full cost of updating the instance population. By knowing all of them in advance, we only consider once each node in the bottom-up pass and once in the top-down pass. We express that dealing with all hiding requests in parallel leads to the minimum number of new instances by:

$$|T_R^P| = \min_i \left| \left(T_{\{i\}}^S\right)_{R-\{i\}}^S \right|$$

The formula states that for a tree $T$, the number of instances ($|T|$), after a parallel ($T^p$) hiding process of all rules (leaves) in $R$, is the optimal along all possible orderings of all serial ($T^s$) hiding requests drawn from $R$. A serial hiding request is implemented by selecting a leaf to be hidden and then, recursively, dealing with the remaining leaves (Lemma 3).

**Lemma 3.** *When serially hiding two non-sibling leaves, the number of new instances to be added to maintain the max:min ratios is larger or equal to the number of instances that would have been added if the hiding requests were handled in parallel.*
(The proof is in (Kalles et al., 2016))

We now demonstrate an example, in which two hiding requests were handled in parallel, using the proposed look ahead technique of linear Diophantine equations. In figure 4, we show the original tree with class distributions of instances across edges.

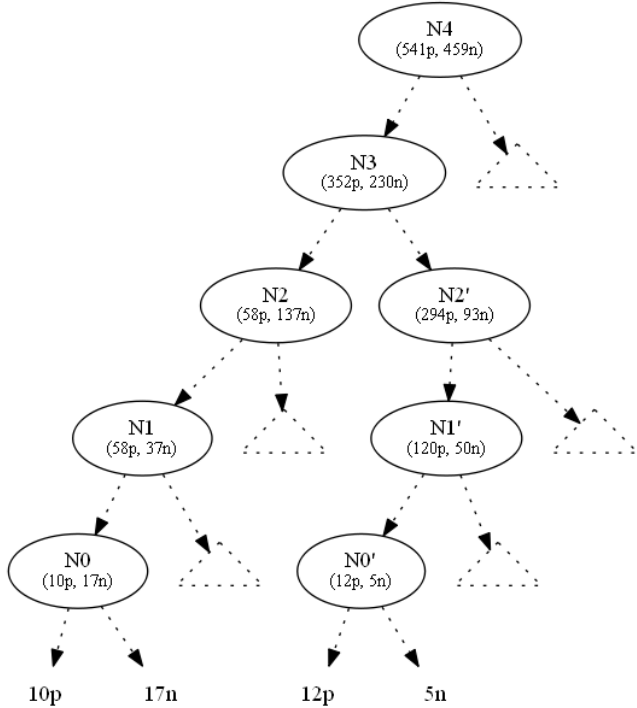

*Figure 4. Original tree*

We use the information gain as the splitting heuristic. To hide the leaf *which corresponds to the 10 positive instances (to the left of N0)* we change the ten positive instances to negative ones and denote this operation by (-10p,+10n). As a result, the parent node, *N0*, becomes a one-class node with minimum (zero) entropy. All nodes located upwards to node *N0* until the root *N4* also absorb the (-10p,+10n) operation (Figure 5).
This conversion would leave *N1* with 48p+47n instances. But, as its initial 58p+37n distribution contributed to *N1*'s splitting attribute, $A_{N1}$, which in turn created *N0* (and then *10p*), we should preserve the information gain of $A_{N1}$, since the entropy of a node only depends on the ratio *p:n* of its instance classes.

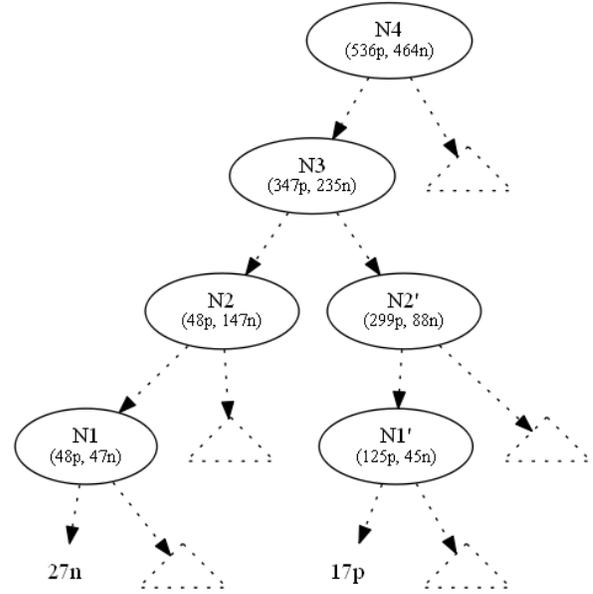

*Figure 5. Bottom-up propagation of instances (-10p,+10n) from the left side and (+5p,-5n) from the right side of the tree.*

To hide the leaf *which corresponds to the 5 negative instances (to the right of N0')* we change the five negative instances to positive ones and denote this operation by (+5p,-5n). As a result, the parent node, *N0'*, becomes a one-class node with minimum (zero) entropy. All nodes located upwards to node *N0'* until the root *N4* also absorb the (-10p,+10n) operation (Figure 5). The intersection node N3 and the root N4 will be affected by (-5p,+5n) which is the total outcome of the two operations from the two subtrees below of N3.

This conversion would leave *N1'* with 125p+45n instances. But, as its initial 120p+50n distribution contributed to *N1*'s splitting attribute, $A_{N1'}$, which in turn created *N0'* (and then *5n*), we should preserve the information gain of $A_{N1'}$, since the entropy of a node only depends on the ratio *p:n* of its instance classes.

In order to maintain the ratio of nodes N1 and N1', we have to add appropriate number of positive and negative instances to N1, N1' and extend this addition process up until the tree root, by accumulating at each node all instance requests from below and by adding instances locally to maintain the node statistics, propagating these changes to the tree root.

Let $(x_1, y_1)$ be the number of positive and negative instances respectively that should be added to node N1 to maintain its initial ratio. This can be expressed with the following equation:

$$\frac{48 + x_1}{47 + y_1} = \frac{58}{37}$$

The above equation is equivalent to the following linear Diophantine equation:
$$37x_1 - 58y_1 = 950 \quad (5)$$
Similarly, let $(x_2, y_2), (x'_1, y'_1), (x'_2, y'_2), (x_3, y_3), (x_4, y_4)$ be the corresponding number of positive and negative instances that should be added to nodes N2, N1', N2', N3 and N4.

The corresponding linear Diophantine equations for nodes N2, N1', N2', N3 and N4 are:
$$137x_2 - 58y_2 = 1950 \quad (6)$$
$$50x'_1 - 120y'_1 = -850 \quad (7)$$
$$93x'_2 - 294y'_2 = -1935 \quad (8)$$
$$230x_3 - 352y_3 = 2910 \quad (9)$$
$$459x_4 - 541y_4 = 5000 \quad (10)$$

The general solutions of the above six (5-10) linear Diophantine equations are given below ( $k \in \mathbb{Z}$ ):

$$37x_1 - 58y_1 = 950 \Leftrightarrow \begin{cases} x_1 = 10450 + 58k \\ y_1 = 6650 + 37k \end{cases}$$
$$137x_2 - 58y_2 = 1950 \Leftrightarrow \begin{cases} x_2 = -21450 + 58k \\ y_2 = -50700 + 137k \end{cases}$$
$$50x'_1 - 120y'_1 = -850 \Leftrightarrow \begin{cases} x'_1 = -425 + 12k \\ y'_1 = -170 + 5k \end{cases}$$
$$137x'_2 - 58y'_2 = 1755 \Leftrightarrow \begin{cases} x'_2 = -12255 + 98k \\ y'_2 = -3870 + 31k \end{cases}$$
$$137x_3 - 352y_3 = 4401 \Leftrightarrow \begin{cases} x_3 = 109125 + 176k \\ y_3 = 71295 + 115k \end{cases}$$
$$459x_4 - 541y_4 = 9000 \Leftrightarrow \begin{cases} x_4 = -165000 + 541k \\ y_4 = -140000 + 459k \end{cases}$$

From the infinite pairs of solutions for every linear Diophantine equation we choose the pairs
$(\overline{x_1}, \overline{y_1}), (\overline{x_2}, \overline{y_2}), (\overline{x'_1}, \overline{y'_1}), (\overline{x'_2}, \overline{y'_2}), (\overline{x_3}, \overline{y_3}), (\overline{x_4}, \overline{y_4})$,
where $\overline{x_1}, \overline{x_2}, \overline{x'_1}, \overline{x'_2}, \overline{x_3}, \overline{x_4}, \overline{y_1}, \overline{y_2}, \overline{y'_1}, \overline{y'_2}, \overline{y_3}, \overline{y_4}$ are the minimum natural numbers that satisfy the conditions (C1) and (C2).
(C1): $\overline{x_1} \leq \overline{x_2}$ and $\overline{y_1} \leq \overline{y_2}$ and $\overline{x'_1} \leq \overline{x'_2}$ and $\overline{y'_1} \leq \overline{y'_2}$
(C2): $\overline{x_2} + \overline{x'_2} \leq \overline{x_3} \leq \overline{x_4}$ and $\overline{y_2} + \overline{y'_2} \leq \overline{y_3} \leq \overline{y_4}$

Condition (C1) ensures that we have selected the optimum path from the leaves up to intersection node N3 of the decision tree.
Condition (C2) ensures that we have selected the optimum path from one level below the intersection node N3 (N2, N2') up to the root.
For this example, the pairs of solutions that are both minimum and satisfy the conditions (C1), (C2) are:
$(\overline{x_1}, \overline{y_1}) = (68, 27)$
$(\overline{x_2}, \overline{y_2}) = (68, 127)$
$(\overline{x'_1}, \overline{y'_1}) = (7, 10)$
$(\overline{x'_2}, \overline{y'_2}) = (93, 36)$
$(\overline{x_3}, \overline{y_3}) = (357, 225)$
$(\overline{x_4}, \overline{y_4}) = (546, 454)$

Based on the above solutions we should add to N1, 68 positive and 27 negative instances. These new instances propagate upwards, therefore on N2, we don't need to add any positive instances but we need to add 100 (=127-27) negative instances. In the same manner, we should add to N1', 7 positive and 10 negative instances. These new instances propagate upwards, therefore on N2', we need to add 86 (=93-7) positive instances and 26 (=36-10) negative instances.

Similarly, for N3 we should add, 196 (=357-68-93) new positive and 62 (=225-127-36) new negative instances. Finally, for N4, we should add 189 (=546-357) new positive and 229 (=454-225) new negative instances.

Therefore, based on this example we observe that by using this technique we can handle more than one hiding requests without any increase in the number of instances that can be added. This proof of concept shows that formulating the problem of hiding requests in parallel is nearly a natural fit for the linear Diophantine equations technique.

## 3 Conclusions and directions for further work

We have presented the outline of a heuristic that allows one to specify which leaves of a decision tree should be hidden and then proceed to judiciously add instances to the original data set so that the next time one tries to build the tree, the to-be-hidden nodes will have disappeared because the instances corresponding to those nodes will have been absorbed by neighboring ones.

We have presented a fully-fledged example of the proposed approach and, along its presentation, discussed a variety of issues that relate to how one might minimize the amount of modifications that are required to perform the requested hiding as well as where some side-effects of this hiding might emerge. To do so, we have turned our attention to using linear Diophantine equations to formulate the constraints which must be satisfied for the heuristic to work.

Of course, several aspects of our technique can be substantially improved.

The *max:min* ratio concept can guarantee the preservation of the information gain of a splitting attribute but it would be interesting to see whether it can be applied to other splitting criteria too. Since this ratio is based on frequencies, it should also work with a similar popular metric, the *Gini* index (Breiman et al., 1984). On the other hand, it is unclear whether it can preserve trees that have been induced using more holistic metrics, such as the minimum description length principle (Quinlan and Rivest, 1989).

Extensive experimentation with several data sets would allow us to estimate the quality of the *max:min* ratio heuristic and also experiment with a revised version of the heuristic, one that strives to keep the *p:n* ratio of a node itself (and not its parent), or one that attempts to remove instances instead of swapping their class labels, or still another that further relaxes the *p:n* ratio concept during the top-down phase by distributing all unspecified instances evenly among the left and right outgoing branch from a node and proceeding recursively to the leaves (which is the one we actually implemented). In general, experimenting with a variety of heuristics to trade off ease of implementation with performance is an obvious priority for experimental research.

On performance aspects, besides speed, one also needs to look at the issue of judging the similarity of the original tree with the one produced after the above procedure has been applied. One might be interested in syntactic similarity (Zantema and Bodlaender, 2000) (comparing the data structures –or parts thereof- themselves) or semantic similarity (comparing against reference data sets). This is an issue of substantial importance, which will also help settle questions of which heuristics work better and which not.

As the number of instances to be added is a main index of the heuristic's quality a reasonable direction for investigation is to determine the appropriate ratio values, which result in smaller integer solutions of the corresponding Linear Diophantine Equations but, at the same time, do not deviate too much from the structure of the original tree. This suggests the adoption of approximate ratios instead of exact ones and, obviously, raises the potential to further investigate the trade-off between data-set increase and tree similarity.

Extensive experimentation with different decision trees would allow us to observe if this look ahead technique can be applied not only to two parallel hiding requests but for any *k* simultaneously specified requests, without any impact to the number of instances that should be added. It should not be ruled out that this could even lead to a formally proven result.

It is rather obvious that the variety of answers one could explore for each of the questions above constitutes a research agenda of both a theoretical and an applied nature. At the same time, it is via extending the base case, by allowing multi-valued and numeric attributes and multi-class problems that we should address the problem of enhancing the basic technique, alongside investigating the robustness of this heuristic to a variety of splitting criteria and to datasets of varying size and complexity. The longer-term goal is to have it operate as a standard data engineering service to accommodate hiding requests, coupled with a suitable environment where one could specify the importance of each hiding request.

We have developed a prototype web-based application which implements the aforementioned technique and we have used it to obtain initial confirmation of the validity of our arguments.